\documentclass[times,referee,twocolumn,final,authoryear]{elsarticle}

\usepackage{ycviu}
\usepackage{framed,multirow}

\usepackage{amssymb}
\usepackage{latexsym}

\usepackage{times}
\usepackage{epsfig}
\usepackage{graphicx}
\usepackage{amsmath}
\usepackage{subfigure}
\usepackage[english]{babel}
\usepackage{booktabs}
\usepackage{lineno}

\usepackage{url}
\usepackage{xcolor}
\definecolor{newcolor}{rgb}{.8,.349,.1}

\def\eg{\emph{e.g. }}

\def\ie{\emph{i.e. }}

\journal{Computer Vision and Image Understanding}

\begin{document}

\thispagestyle{empty}

\ifpreprint
  \setcounter{page}{1}
\else
  \setcounter{page}{1}
\fi

\begin{frontmatter}

\title{Unsupervised learning from videos using temporal coherency deep networks}

\author[1]{Carolina \snm{Redondo-Cabrera}\corref{cor1}} 
\cortext[cor1]{Corresponding author: 
  Tel.: +34-91-885-6732;  
  fax: +34-91-885-6699;}
\ead{carolina.redondoc@edu.uah.es}
\author[1]{Roberto \snm{Lopez-Sastre}}

\address[1]{GRAM, University of Alcal\'a, Alcal\'a de Henares, 28805, Spain}

\received{1 May 2013}
\finalform{10 May 2013}
\accepted{13 May 2013}
\availableonline{15 May 2013}
\communicated{S. Sarkar}
\begin{abstract}
In this work we address the challenging problem of unsupervised learning from videos. Existing methods utilize the spatio-temporal continuity in contiguous video frames as regularization for the learning process. Typically, this temporal coherence of close frames is used as a free form of annotation, encouraging the learned representations to exhibit small differences between these frames. But this type of approach fails to capture the dissimilarity between videos with different content, hence learning less discriminative features. We here propose two Siamese architectures for Convolutional Neural Networks, and their corresponding novel loss functions, to learn from unlabeled videos, which jointly exploit the local temporal coherence between contiguous frames, and a global discriminative margin used to separate representations of different videos. An extensive experimental evaluation is presented, where we validate the proposed models on various tasks. First, we show how the learned features can be used to discover actions and scenes in video collections. Second, we show the benefits of such an unsupervised learning from just unlabeled videos, which can be directly used as a prior for the supervised recognition tasks of actions and objects in images, where our results further show that our features can even surpass a traditional and heavily supervised pre-training plus fine-tunning strategy.
\end{abstract}

\end{frontmatter}


\vspace{-0.6cm}
\section{Introduction}
\label{sec:intro}
Numbers speak for themselves: more than 300 hours of video content are added to YouTube every minute\footnote{\url{https://www.youtube.com/yt/press/statistics.html}}. A need that certainly arises from such huge source of on-line videos is the ability to automatically learn from it, ideally without requiring any annotation or form of supervision. 

State of the art approaches learning from videos are \emph{fully supervised}, \eg \citep{ji2013,Karpathy2014,Simonyan2014,Herath2017}. This means they all rely on large-scale video datasets heavily annotated, an aspect that has not been a problem so far: every year a novel fully annotated video dataset appears (\eg HMDB51 \citep{Kuehne2011}, UCF101 \citep{Soomro2012}, THUMOS \citep{Idrees2017},  ActivityNet \citep{Caba-Heilbron2015}). 

So, the question is: why unsupervised learning? The first answer is obvious: the use of a supervised model implies a tremendous effort annotating data, a time-consuming task, with its associated costs, and that is irremediably prone to user-specific biases and errors. But there are more reasons that justify the need of a truly unsupervised learning approach, especially for videos. 

Fundamentally, videos are much higher dimensional entities compared to images. Therefore, to learn long range structures generally implies to do a lot of feature engineering, \eg computing the right kind of flow and trajectories features \citep{Wang2015,Turchini2017}, or using elaborated feature pooling methods \citep{Fernando2016}. Furthermore, like we pointed above, every minute hundreds of videos are freely available without annotations, and none of these heavily supervised solutions can benefit from this invaluable source of information.

Current unsupervised methods \citep{Jayaraman2016,Misra2016,Mobahi2009,Srivastava2015} use the temporal coherence of contiguous video frames as regularization for the learning process. This way, the spatio-temporal continuity of close frames is used as a free form of annotation or supervision, which is used to learn representations that exhibit small differences between these frames. However, these approaches fail to capture the dissimilarity between videos with different content, hence learning less discriminative features.

\begin{figure}[t]
\centering
\includegraphics[width=\linewidth]{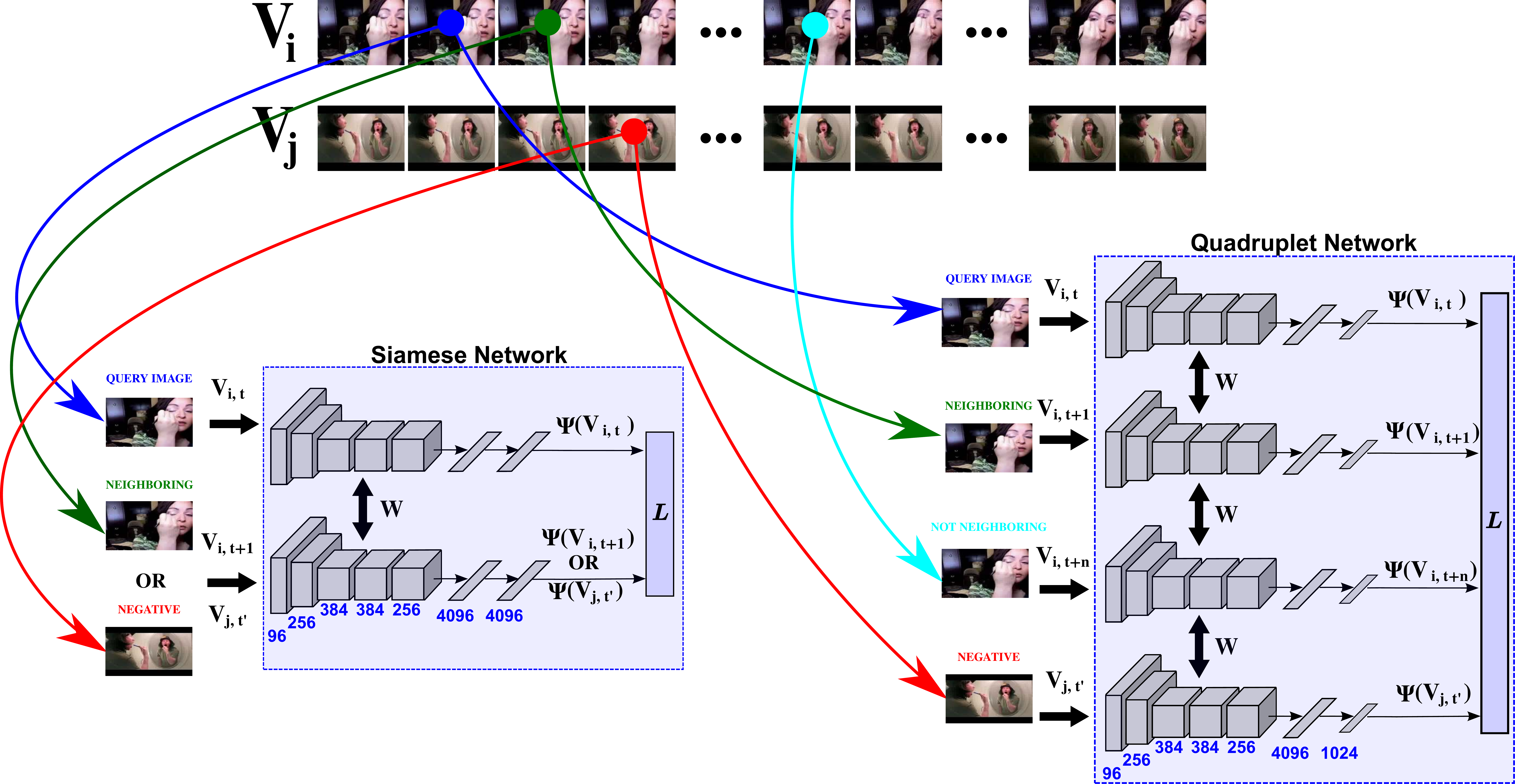}
\caption{Siamese deep learning network configurations for unsupervised learning from videos. Our models simultaneously exploit the local slowness present in a video sequence, and a global discriminative margin between two different video sequences.}
\label{fig:network}
\end{figure}

In this paper, we also choose to work in a fully unsupervised learning setting, where our models only have access to unlabeled videos. The key contributions of our work can be summarized as follows:

\begin{itemize}
\item We introduce two novel fully unsupervised learning approaches, based on Siamese Convolutional Neural Network (CNN) architectures, which exploit the local temporal coherence between contiguous frames, while learn a discriminative representation to separate videos with different content by a margin. Our models and their associated novel loss functions are detailed in Section \ref{sec:approach}. Figure \ref{fig:network} shows a graphical description of our approach. 
\item In order to demonstrate the effectiveness of our features, we propose to address the problems of unsupervised discovery of actions and scenes in video collections (see Section \ref{sec:discovering}). In other words, given a set of unlabeled videos, the goal is to separate the different classes of actions or scenes. To the best of our knowledge, we propose these two novel experiments, with their two rigorous frameworks, where the objective is to evaluate how the models discover the different actions and scenes, using clustering validation metrics, according to a set of categories. These experimental validations are presented in Sections \ref{sec:results_action} and \ref{sec:results_scenes}, for actions and scenes, respectively.

\item Finally, we further evaluate our unsupervised learning approaches, using them as the initialization process for different deep learning architectures, which are then fine-tuned to address the supervised tasks of: a) human action recognition (Section \ref{action_recognition}); and b) object recognition (Section \ref{object_recognition}). Very interestingly, our results show that our initialization strategy can even surpass the standard and heavily supervised pre-training plus fine-tunning approach.
\end{itemize}

\section{Related Work}
\label{sec:relatedwork}
Although there have been models for unsupervised feature learning from images, \eg \citep{Kihyuk2012,Dosovitskiy2015}, we focus here on the different problem of unsupervised feature learning using videos. The first approaches for unsupervised feature learning from videos were based on dimensionality reduction techniques, \eg \citep{Hateren1998, Hurri2003}. But, certainly, most of prior work leveraging video for unsupervised feature learning relies on the concept of slow feature analysis (SFA) \citep{Wiskott2002}. Essentially, this idea proposes to exploit the temporal coherence in consecutive video frames as a form of free supervision to learn the visual representations. Although Long Short-Term Memory (LSTM) based approaches have been proposed, \eg \citep{Srivastava2015}, most recent work combines CNNs with the SFA strategy \citep{Jayaraman2016,Misra2016,Mobahi2009}. \cite{Mobahi2009} introduce a regularizer to encourage this coherence using a contrastive hinge loss.  \cite{Misra2016} propose to shuffle a set of consecutive frames and train a set of Siamese deep networks to detect whether a sequence of frames is in the correct temporal order. In \citep{Jayaraman2016}, the feature analysis is generalized from \emph{slow} to \emph{steady}, proposing a model that requires not only that sequential features change slowly, but also that they change in a similar manner in adjacent time intervals of the video sequence. Furthermore, the objective function in \citep{Jayaraman2016} combines a supervised loss with an unsupervised regularization term, simultaneously. 

SFA-based learning has been also used for the problem of dimensionality reduction \citep{Hadsell2006}, or for action recognition \citep{Sun2014}. Finally, other approaches learn a specific representation for tracked local patches \citep{Wang2015a, Zou2011}, and apply the features to the object recognition problem.

Related to the above mentioned methods, our aim is also to learn features from unlabeled videos. However, whereas all the past work focuses on preserving \emph{only} this temporal coherence among near frames \citep{Jayaraman2016,Misra2016,Mobahi2009}, our learning objective is the first to move beyond this slowness, requiring also that frames extracted from different randomly selected videos do not lie close in the feature space to be learned. Technically, we propose two deep learning models, a Siamese and a Quadruplet CNN architecture, which learn a feature space where the temporal coherence in a video sequence is kept. But, simultaneously, they enforce a representation where the frames extracted from the same video should be much closer than any other random pair of frames extracted from different videos. Note that our approach is purely unsupervised, so no combination with any form of supervision in the objective function is allowed, in contrast to \citep{Jayaraman2016}. Also, the additional steadiness enforced in \citep{Jayaraman2016} does not guarantee any margin between visual representations for different videos. \cite{Wang2015a} also propose to use triplets where the negative frame is extracted from a different video, however the differences of our techniques with respect to \citep{Wang2015a} are notable. First, they propose to work with tracked image patches, instead of frames. Their model requires of an extra patch extraction step using a tracking process, based on matching SURF \citep{Bay2008} features and the Improved Trajectories \citep{Wang2013} techinique. We simply propose to use the video frames, learning the model completely end-to-end. 

Finally, our model extends the typical triplet Siamese network architecture \citep{Jayaraman2016,Wang2015a} to a quadruplet Siamese network. Our novel architecture allows to enrich the traditional SFA methodology, extending the learning based on temporal coherence from \emph{between frames in the same video} to \emph{between frames in different videos}. This way, we introduce a novel layer of regularization which has not been previously explored.

Technically, we provide two approaches to directly learn an embedding into an Euclidean space for video features. Therefore, we can consider our models as metric learning solutions. Our Siamese and Quadruplet deep learning architectures are learned using some modifications proposed for the contrastive loss \citep{Hadsell2006} using triplets and quadruplets of features. Similar triplet loss functions have been proposed for the problem of face verification \citep{Schroff2015}, image ranking \citep{Wang2014}, or for learning local descriptors \citep{Simo-Serra2015,Balntas2016,KumarBG2016}. Some of these models propose to include a hard negative mining strategy, which in our case has not been implemented. The main reason is that the definition of a hard negative itself in our unsupervised setting for the videos is ambiguous: should we prefer to identify hard negative frames from the same videos used to extract the contiguous frames, or hard negative examples that come from different videos?. Moreover, \cite{KumarBG2016} propose the combination of the triplet loss with a global loss to minimize the overall classification error in the training set. On the contrary, our model is completely unsupervised, hence the combination with a global classification loss is not possible, either.

\section{Unsupervised Learning from Videos}
\label{sec:approach}

Technically, our problem consists in training a CNN for feature learning, given a set of $N$ unlabeled videos $\mathcal{S}=\{V_1, V_2, \ldots, V_N\}$.  Being $W$ the parameters of the network, our objective function can be expressed as,

\begin{equation}
\label{obj}
\min\limits_{W} \frac{\lambda}{2} ||W||^2 + \sum_{i=1}^{T} \mathcal{L}_u(W,\mathcal{U}_i) \, ,
\end{equation}
where $\lambda$ is the weight decay constant, $\mathcal{U}_i$ encodes the set of $T$ training tuples of video frames, and $\mathcal{L}_u$ represents the unsupervised regularization loss term.

In the following, we introduce two novel Siamese deep learning architectures, and design two loss functions for $\mathcal{L}_u$ to train the networks in an unsupervised way.

\subsection{The Siamese Architecture}

According to the SFA paradigm introduced by \cite{Wiskott2002}, it is possible to learn from slowly varying features from a vectorial input signal. Videos become then ideal candidates as input signals. We simply have to comprehend that the temporal coherence between contiguous video frames can work as a form of supervision, which is free and structural.

SFA encourages the following property: in a learned feature space, temporally nearby video frames should lie close to each other, \ie for a learned representation $\Psi$, and adjacent video frames $V_{i,t}$ and $V_{i,t+1}$, one would like $\Psi(W,V_{i,t}) \approx \Psi(W,V_{i,t+1})$. In the following, we will abbreviate $\Psi(W,V_{i,t})$ to $\Psi(V_{i,t})$, assuming the learned feature representation is a function of the learned parameters $W$ for the network.

The key idea behind our first Siamese model is as follows. During learning, we want to encourage that a query frame and its adjacent video frame lie close in the learned feature space $\Psi$, \emph{while} the distance between this query frame and any other randomly extracted frame coming from a \emph{different video} is higher than a margin ($\delta$). 

So, we first propose the Siamese architecture shown in Figure \ref{fig:network} (left). It consists of two base networks which share the same parameters. The input of the model is a pair of video frames, $(V_{i,t},V_{i,t+1})$ or $(V_{i,t},V_{j,t'})$, and an associated label $Y$ ($Y = 1$ when the frames belong to the same video, $Y = 0$ otherwise). The final output of each base network is a 4096-dimensional feature space $\Psi(\cdot)$. Note that we do not initialize this base network with weights from a previous training in any other dataset, like, for instance, ImageNet. Technically, we start with a random initialization for the model parameters. So, our approach is \emph{fully} unsupervised.

We want our network to learn a feature representation $\Psi(\cdot)$, such that, for a query video $V_i$ and a randomly chosen different video $V_j$, the distance between the feature representations of the neighboring frames $V_{i,t}$ and $V_{i,t+1}$ is lower than the distance from our query frame $V_{i,t}$ and any other frame randomly extracted from the video $V_j$, \ie $V_{j,t^{'}}$.

For doing so, we propose the following contrastive loss function, based on \citep{Hadsell2006}, which can also exploit negative (non-neighbor) pairs,
\begin{eqnarray}
\begin{split}
\label{eq:loss_cont_2}
\mathcal{L}_s(\Psi(V_{x}), \Psi(V_{y}))= & & \\
d(\Psi(V_{x}), \Psi(V_{y})), &  & \textnormal{if } V{x} = V_{i,t} \textnormal{ and } V{y} = V_{i, t+1}\\
\max\{0, \delta- d(\Psi(V_{x}), \Psi(V_{y})) & & \textnormal{if } V{x} = V_{i,t} \textnormal{ and } V{y} = V_{j, t^{'}} \,.
\end{split}
\end{eqnarray}


For the distance function $d$ we follow the euclidean distance. As it is shown in Equation \ref{eq:loss_cont_2}, our contrastive loss penalizes the distance between pair of neighbor frames, and encourages the distance between them when they do not belong to the same video. 

We do not use any video labels during training, therefore, it could happen that the randomly selected different video, \ie $V_{j}$, might belong to the same category of $V_{i}$. If one plans to use the learned features for classifications tasks, as we do, might this be problematic? We assume that there is little likelihood that this situation actually happens, but in case it does, we consider that to impose the margin between frames of different videos, even if they belong to the same category, pushes the feature learning towards more fine-grained representations, able to capture higher-level semantic visual concepts.

\subsection{The Quadruplet Siamese Architecture}

With our previous Siamese architecture, the model is not able to \emph{see} how the learned feature representation evolves within the same video. Note that only contiguous frames are used. How can we further capture rich structure in how the visual content changes over time?

\cite{Jayaraman2016} propose to generalize the SFA model in \citep{Wiskott2002}, introducing a feature learning model which is able to impose a second order temporal derivative during learning the video representation. Using triplets of temporally close frames ($V_{i,a},V_{i,b},V_{i,c}$), their model guarantees that feature changes in the immediate future remain similar to those in the recent past, \ie $d(\Psi(V_{i,b}),\Psi(V_{i,a})) \approx d(\Psi(V_{i,c}),  \Psi(V_{i,b}))$. 

The key idea of our approach is different. We want to guarantee that in the learned feature representation the temporal coherence of contiguous frames is kept, \emph{while} the distance between frames in a wider temporal locality, defined by a temporal window of length $n$, is always lower than the distance between frames coming from different videos. In summary, we propose a model able to capture the local slowness of the video, keeping a global \emph{discriminative} representation. Furthermore, while the objective function in \citep{Jayaraman2016} combines a supervised loss with an unsupervised regularization term, our model is completely unsupervised.

For doing so, we propose the Quadruplet Siamese architecture illustrated in Figure \ref{fig:network} (right). It consists of four base networks which share the same parameters. The input of the model corresponds to the tuple of four video frames $(V_{i,t},V_{i,t+1},V_{i,t+n},V_{j,t'})$. The output of each base network is now a 1024-dimensional feature space.

\begin{figure}[t]
\centering
\includegraphics[width=\linewidth]{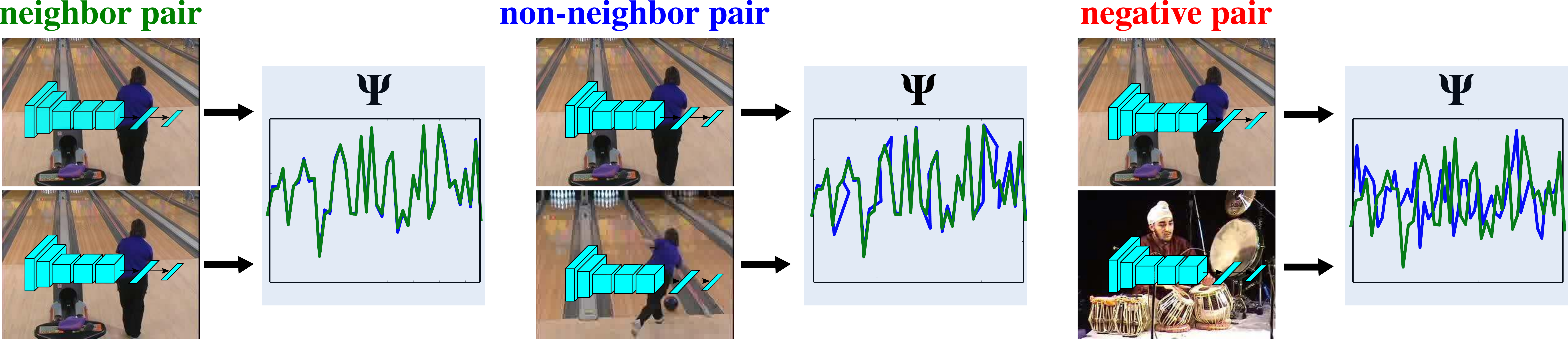}
\caption{Quadruplet Siamese network key idea: A query frame and its adjacent frame should lie close in the learned feature space, while this query frame share a more similar representation with a non-neighbor frame than with a frame extracted from a different video. The feature representation ($\Psi$) for the query frame is shown in green, and the representations for the neighbor, the non-neighbor and the negative frames appear in blue. Note that, when the frames belong to the same video the blue and green representations are closer than when the frames belong to different videos. Consequently, the proposed architecture is able to learn an embedding of video frames into a feature space, where some semantic is preserved, \eg frames with similar/dissimilar ``actions'' or ``scenes'' are mapped to close/far points in the discovered space.}
\label{fig:key_idea}
\end{figure}

Why these four video frames? As it is shown in Figure \ref{fig:key_idea}, we want that a query frame $V_{i,t}$ and its adjacent video frame $V_{i,t+1}$ lie close in the learned feature space ($\Psi(V_{i,t}) \approx \Psi(V_{i,t+1})$), while this query frame shares a more similar representation with a non-neighbor frame $V_{i,t+n}$ of the same video, than with a a frame extracted from a different video $V_{j,t'}$, \ie $d(\Psi(V_{i,t}),\Psi(V_{i,t+n})) < d(\Psi(V_{i,t}),\Psi(V_{j,t'}))$.

This is encoded in an improved loss function, designed to train the proposed Quadruplet architecture, as follows:

\begin{equation} \begin{split}
 \label{loss}
\mathcal{L}_{q}(\Psi(V_{i,t}), \Psi(V_{i,t+1}), \Psi(V_{i,t+n}), \Psi(V_{j,t^{'}}))= & \\
d(\Psi(V_{i,t}), \Psi(V_{i,t+1})) + & \\ 
\max\{0, d(\Psi(V_{i,t}), \Psi(V_{i,t+n})) - d(\Psi(V_{i,t}), \Psi(V_{j,t^{'}})) + \alpha  \},
\end{split}\end{equation}
where $\alpha$ represents our \textit{discriminative} global margin.

\subsection{Implementation details} 

For all our unsupervised learning approaches, we apply mini-batch Stochastic gradient descent (SGD) during training. Our solutions have been built using the Caffe framework \citep{jia2014caffe}. Input video frames are scaled to a fixed size of $227\times227$ pixels. The base network of all our architectures follows the AlexNet model of \cite{Krizhevsky2012} for the convolutional layers. Then we stack two fully connected layers on the pool5 layer outputs, whose neuron numbers are: a) 4096 and 1024 for the Quadruplet network trained with $\mathcal{L}_q$; and b) 4096 and 4096 for the Siamese model trained with $\mathcal{L}_s$. All our architectures share the parameters, therefore we can perform the forward propagation for the whole batch by a single base network, and calculate the loss based on the output features. To train our models, we set the batch size to $120$ pairs and $40$ tuples of frames, for the Siamese and Quadruplet networks, respectively. The learning rate starts with $\epsilon_0 = 0.001$ for both deep architectures. For the Siamese network $\delta = 1$, and for the Quadruplet network the global discriminative margin, $\alpha$, is fixed to $0.5$, and $n=20$. All these parameters have been cross-validated. Experiments show all the details.

\section{Discovering Actions and Scenes in Videos from A to Z}
\label{sec:discovering}
\begin{figure}[t]
\centering
\includegraphics[width=\linewidth]{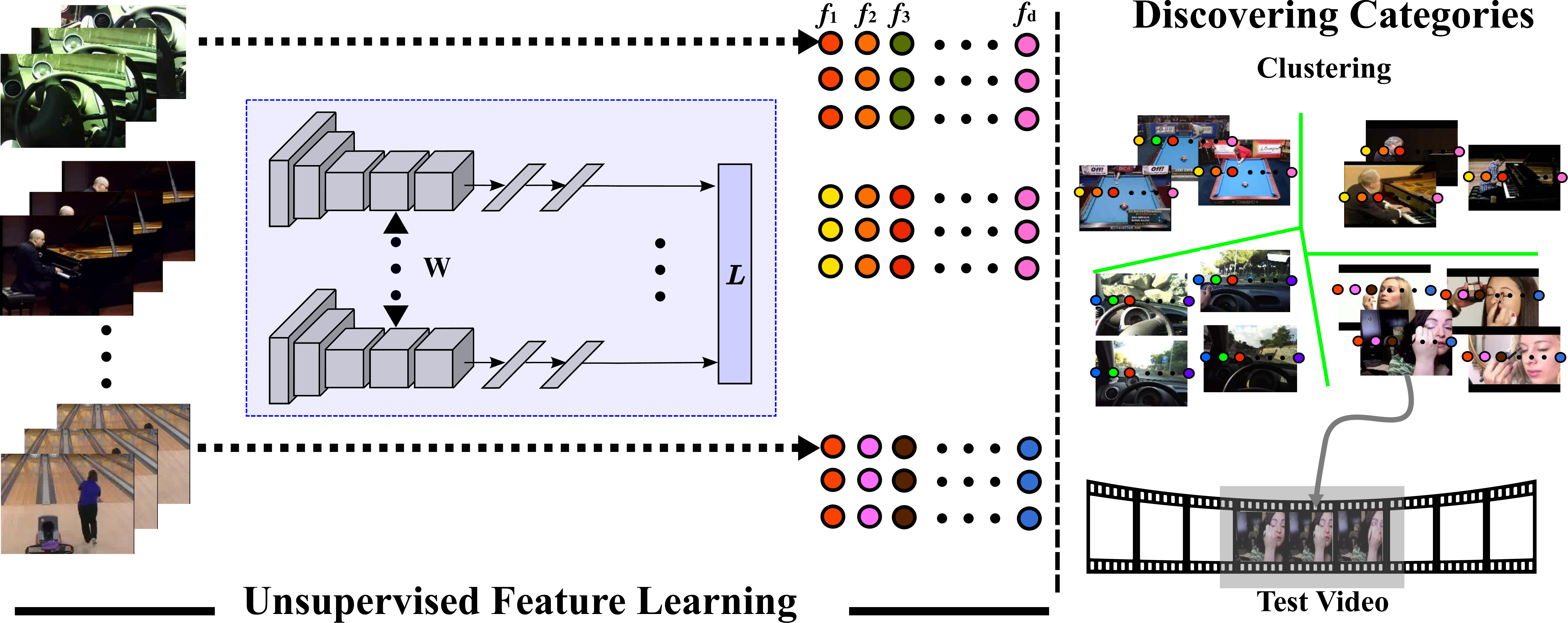}
\caption{Unsupervised action and scene discovery in videos. Our Siamese architectures learn, in a fully unsupervised way, the features to be used for the discovery task in videos. The different categories can be discovered running a clustering algorithm over the features.}
\label{fig:approach_global}
\end{figure}

Figure \ref{fig:approach_global} illustrates the designed approach  for discovering actions or scenes in videos. Given a set of $N$ unlabeled videos $\mathcal{S}=\{V_1, V_2, \ldots, V_N\}$, our goal is to separate the different actions or scenes they represent. Here we take inspiration from the object discovery approach in still images, introduced by \cite{Tuytelaars2010}. Essentially, we generalize their evaluation procedure and problem description, from the discovery of objects in image collections, to the unsupervised discovery of actions and scenes in videos.

We consider that this is an ideal fully unsupervised scenario, for an adequate experimental validation of the unsupervised feature learning approaches that we introduce in this paper.

In particular, we design two experiments with two video datasets: one for discovering scenes, the 5-Contexts dataset \citep{Furnari2015}; and the UCF101 \citep{Soomro2012} for the discovery of actions. More details of the experimental setups are given in Section \ref{sec:exp_setup_discovery}.

As it is shown in Figure \ref{fig:approach_global}, we consider that each video frame $V_{i,t}$ can be characterized by a feature vector $\Psi(V_{i,t})$ which is learned by one of our architectures in an unsupervised way. With all these feature vectors, for all the frames of all the videos, we can proceed to discover the different scenes or actions.  Because the state of the art for discovering objects in image collections has been reached by clustering based methods \citep{Tuytelaars2010}, we propose to solve our described tasks with a clustering algorithm over the learned features. Note that this pipeline can be generalized to any other partition technique.

Built the partitions, it is time to evaluate the performance of the discovery tasks. Technically, the video frames have been separated into a set of clusters. Therefore, as in \citep{Tuytelaars2010}, we propose to use standard metrics for scoring the discovery quality when clusters are used, in particular the conditional entropy.

For the conditional entropy (CE), we follow the classical formulation proposed in Information Theory. Consider the ground truth category labels $X$ and obtained cluster labels $Y$. So, given variables $(x,y)$, sampled from the finite discrete joint space $X \times Y$, we define CE as follows,

\begin{equation}
H(X|Y) = \sum_{y \in Y} p(y) \sum_{x \in X} p(x|y) log\left(\frac{1}{p(x|y)}\right) \, .
\end{equation}

In a nutshell, our experiments consists of the following steps. First, a set of training videos are used to learn the proposed architectures in an unsupervised way, see Section \ref{sec:approach}. Second, we proceed to compute, for the test frames of the test videos, their corresponding feature representation. Once these test features have been obtained, we run a clustering algorithm and evaluate its discovery performance. Note that as in \citep{Tuytelaars2010}, at this point we must fix the number of clusters to be build by the clustering algorithms. Ultimately, the machine should be able to decide on this too, without any human intervention, but here we want to define a clear experimental evaluation where the different models can be compared given a known number of classes. Note that we do not perform any parameter tuning. Each experiment is simply repeated 10 times, and its average performance and standard deviation are reported, according to the CE metric.

\subsection{Experiments}

We have designed these experiments to accomplish the following targets: a) get a quantitative and qualitative understanding of how our unsupervised models discover scenes and actions in videos; and b) evaluate whether the learned representations are able to generalize well, when the problem changes.

\subsubsection{Experimental Setup} 
\label{sec:exp_setup_discovery}

To discover actions in videos, we propose to use the UCF101 dataset \citep{Soomro2012}. It is an action recognition dataset of realistic action videos, collected from YouTube. It consists of over $12.000$ videos categorized into 101 human action classes. The dataset is divided in three splits. For our experiments we use the split-1, which consists of 9537 videos for training and 3783 for testing.

For discovering scenes in videos, we have chosen the publicly available 5-Contexts dataset released by \cite{Furnari2015}. It consists of multiple egocentric videos acquired by a single user in different personal contexts (car, coffee vending machine, office, TV and home office). Four different wearable cameras are used for recording the scenes, assuming that the user can turn his head and move his body when interacting with the environment. We follow the training/test splits provided with the dataset. The training set consists of $5\times4$ videos of 10 seconds. For the test set, the dataset provides medium length videos (from 8 to 10 minutes) with a normal activity of the user in the corresponding personal context.

Although all these datasets come with annotations for all the videos, note that we do \emph{not} use the video annotations provided for learning our features. Instead, we train our unsupervised models with a subset of random video frames extracted from the videos. Technically, our models are learned collecting pairs and tuple of frames, randomly sampled from the training videos.

\subsubsection{Discovering Actions}
\label{sec:results_action}

Let us use the challenging UCF101 dataset to offer a detailed analysis of the proposed models. We start proposing a pipeline to perform the discovery of action categories, where the performance of our Siamese and Quadruplet architectures is evaluated using two different clustering algorithms over the learned features: $K$-means ($K$ = 101) and Spectral clustering (SC). For the SC we follow the Normalized Cuts Spectral Clustering of \cite{Ng2001}, using a Gaussian kernel for the similarity matrix. 

\begin{figure}[t]
\centering
\includegraphics[trim={110 220 120 220},clip, width=0.8\linewidth]{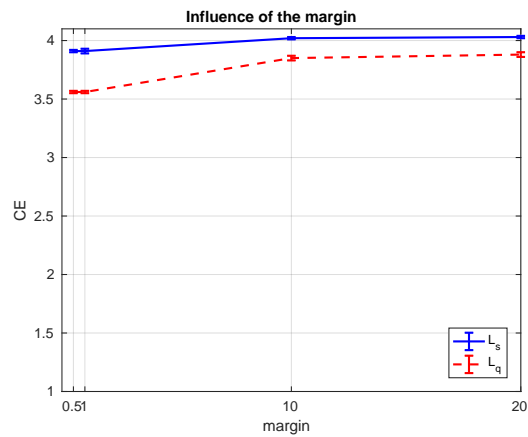}
\caption{Influence of the margin in the architectures proposed.}
\label{fig:parameters_analysis}
\end{figure}

We start analyzing the influence of the model parameters. All our parameters have been cross validated using this dataset. We here detail the influence of: a) the margins ($\delta$ for $\mathcal{L}_s$  and $\alpha$ for $\mathcal{L}_q$);
b) the temporal window to consider contiguous frames ($w$); and the non-neighbor frame index $n$ of $\mathcal{L}_q$. Figure \ref{fig:parameters_analysis} shows how the CE changes when we try with different margins. While the $\mathcal{L}_s$ prefers a margin of 1, $\mathcal{L}_q$ reports the best performance for $\alpha=0.5$. It is worth noting that for both losses, CE increases if the margin is greater than 10. For the $\mathcal{L}_s$, the influence of $w$ is insignificant, as it is also the case with the parameter $n$ for the $\mathcal{L}_q$ loss. The CE reported varying these two parameters remains very similar in all our experiments, considering $w \in [1,10]$ and $n \in [10,30]$. We can conclude that the best combinations of parameters are: 1) for $\mathcal{L}_s$,  $\delta = 1$ and $w=1$ ; 2) for $\mathcal{L}_q$, $\alpha=0.5$, $w=1$ and $n=20$. We fix these values for the rest of the paper.

We now investigate the impact of learned features obtained from different layers of our CNNs, \ie fc6 and fc7 layers. Results are summarized in Table \ref{table:ce_results}.

From Table \ref{table:ce_results}, the comparison between the Siamese architecture and the Quadruplet model reveals that the latter is always better. It is also interesting to observe that the features that best performance are offering come from layer fc6 instead of from fc7, for all the clustering methods and architectures. Finally, results show that the SC is able to discover slightly better categories.

\begin{table}[t]
\caption{Comparison of the different unsupervised approaches using $K$-means and SC in the UCF101 dataset. CE (lower is better) metric is reported.}
\label{table:ce_results}
\begin{center}
\scalebox{0.8}{
\begin{tabular}{lcc}
\toprule
Methods & $K$-means & SC \\
\midrule
AlexNet with $\mathcal{L}_s$ fc6 & 3.91 $\pm$ 0.02 & 3.75 $\pm$ 0.01 \\
AlexNet with $\mathcal{L}_s$ fc7 & 3.95 $\pm$ 0.02 & 3.78 $\pm$ 0.01\\
\midrule
AlexNet with $\mathcal{L}_q$ fc6 & \textbf{3.56 $\pm$ 0.01}  & \textbf{3.39 $\pm$ 0.01} \\
AlexNet with $\mathcal{L}_q$ fc7 & 3.61$\pm$ 0.01  & 3.41 $\pm$ 0.02\\
\midrule
\midrule
AlexNet with $\mathcal{L}_s$ fc6 MU& 3.93 $\pm$ 0.01  & 3.78 $\pm$ 0.01 \\
AlexNet with $\mathcal{L}_s$ fc7 MU& 3.95 $\pm$ 0.02  & 3.79 $\pm$ 0.01 \\
\midrule
AlexNet with $\mathcal{L}_q$ fc6 MU& 3.51 $\pm$ 0.01  & 3.35 $\pm$ 0.01 \\
AlexNet with $\mathcal{L}_q$ fc7 MU& 3.53 $\pm$ 0.01  & 3.38 $\pm$ 0.01\\

\bottomrule
\end{tabular}
}
\end{center}
\end{table}

\textbf{What if we move towards a more unsupervised setup?} The videos of the UCF101 dataset are \emph{trimmed}, \ie they contain the information of just one action. Therefore, our models implicitly use this form of supervision during learning. Although the negative frame can be extracted from a video with the same action category as for the video in which we extract the neighbor frames, we want to explore here a \emph{more unsupervised} scenario. We hence propose a novel experiment, where a unique \emph{long} video is used for learning the features. This \emph{long} video is built concatenating in a random order, all the videos in the UCF101 dataset. Now, the \emph{negative} frame is simply randomly chosen from the long video. Results are reported in Table \ref{table:ce_results}, last 4 rows, using the suffix MU, which stands for More Unsupervised. Overall, the experiment shows that there is no noticeable difference when we follow this more unsupervised procedure. $\mathcal{L}_q$ is still better than  $\mathcal{L}_s$. Interestingly, $\mathcal{L}_q$ benefits from this increased unsupervision, slightly reducing its conditional entropy. As a conclusion, we believe that this aspect reveals the validity of the designed unsupervised experimental validation.

We now offer in Table \ref{table:ce_results_baselines} a comparison of our models with some competing methods. The first one consists in a random assignment of video frames to clusters (Baseline RANDOM in Table \ref{table:ce_results_baselines} -- first row). Baseline RANDOM provides a sanity check in that other methods should always perform better, as it is the case. Note that randomly assigning video frames to clusters results in a CE of $6.31$. This value is close to the maximum $log_{2}(101) = 6.66$, giving an idea of the difficulty of the action discovery task proposed for this dataset.

Another competing method we propose here follows the standard SFA implementation based on a contrastive loss \citep{Hadsell2006}. Technically, the baseline implemented is identical to our Siamese architecture with AlexNet, but instead of using our $\mathcal{L}_s$ loss, a contrastive loss is used with contiguous and not-contiguous frames coming from the videos, following \citep{Mobahi2009}. We consider this comparison important, because it is a form of ablation study, where we empirically validate the benefits of using our novel loss functions, which incorporate the concept of global margin between different videos.

\begin{table}[t]
\caption{Unsupervised discovery of actions in the the UCF101 dataset. Comparison with baselines. CE (lower is better) metric is reported.}
\label{table:ce_results_baselines}
\begin{center}
\scalebox{0.7}{
\begin{tabular}{lcc}
\toprule
Methods & $K$-means & SC\\
\midrule
Baseline RANDOM & 6.31 $\pm$ 0.00 & 6.31 $\pm$ 0.00\\ %
\midrule
AlexNet with $\mathcal{L}_{SFA}$ fc6  & 3.75 $\pm$ 0.03 & 3.55 $\pm$ 0.03\\
\midrule
AlexNet with our $\mathcal{L}_s$ fc6 & 3.91 $\pm$ 0.02 & 3.75 $\pm$ 0.02\\
AlexNet with our $\mathcal{L}_q$ fc6 & \textbf{3.56 $\pm$ 0.01}  & \textbf{3.39 $\pm$ 0.01} \\
\midrule
Pretrained VGG of \citep{Wang2017} fc6 & 3.38 $\pm$ 0.03  & 3.12 $\pm$ 0.02 \\
\bottomrule
\end{tabular}
}
\end{center}
\end{table}

Only the Siamese model features are not able to outperform in this dataset the performance of the SFA approach. However, the best performing method is our Quadruplet network, which clearly outperforms the SFA baseline. This justifies the necessity of incorporating into the learning model not only the global margin for features from different videos, but also the capability of our model to measure how the content changes over time within the same video. That is, the global margin is guaranteed, but letting the temporal locality of the features in the same video to play a role. Overall, by applying our unsupervised procedure with SC, the remaining uncertainty on the true action category has been reduced from $2^{6.31} = 79.34$ for the random assignment, down to $2^{3.39} = 10.48$ for our Quadruplet network.

Table \ref{table:ce_results_baselines} finally shows a direct comparison with the recent work of \cite{Wang2017}. For this experiment, we simply have decided to evaluate the performance of the features learned following \cite{Wang2017} in our proposed problem for action discovery. For doing so, we proceed to use their publicly available pre-trained model to extract the features in our UCF101 experiment. Their CE is slightly better than ours. Note that their model follows a deeper VGG \citep{Simonyan2015} architecture, and that a very different training procedure is used. First, \cite{Wang2017} use more than 100K videos. Then, they need to mine the moving objects in the videos, and to define the inter-instance and intra-instance relations between patches, to build their graph based approach.

\textbf{How does the architecture affect the results?} We here propose to change the AlexNet base network of our approaches, by the more modern GoogleNet, which also has less parameters. We use this network with a fc layer of size 1024 for the feature extraction.
The CE reported for both $\mathcal{L}_s$ and $\mathcal{L}_q$ is of 3.93 $\pm$ 0.01 and of 3.93 $\pm$ 0.02, respectively. First, we observe that now there is no difference between the two loss proposed. Second, overall, the CE increases with this novel architecture. As a conclusion, we can affirm that while supervised methods tend to improve substantially with newer architectures that have more capacity, this is not the case for our unsupervised feature learning proposed models. The overparameterize AlexNet network is the one learning the best representations for the proposed problems.

\textbf{Are our unsupervised models able to compete with standard supervised approaches used for learning the features?} We can now compare the performance of our fully unsupervised solutions, with the performance reported when the features  are extracted using a supervised AlexNet model trained on ImageNet. The CE reported by K-means and SC with fc6 layer features obtained from this supervised model is of 2.68 $\pm$ 0.42 and 2.50 $\pm$ 0.04, respectively. Our best Quadruplet network, trained with roughly 9.500 videos in a fully unsupervised way, reports a CE of 3.39 $\pm$ 0.01. This means that our unsupervised solution, which does not need need any annotation, is able to recover $72\%$ of the supervised performance. 

\begin{figure}[t]
\subfigure[AlexNet trained with $\mathcal{L}_s$]{
\includegraphics[width=0.9\linewidth]{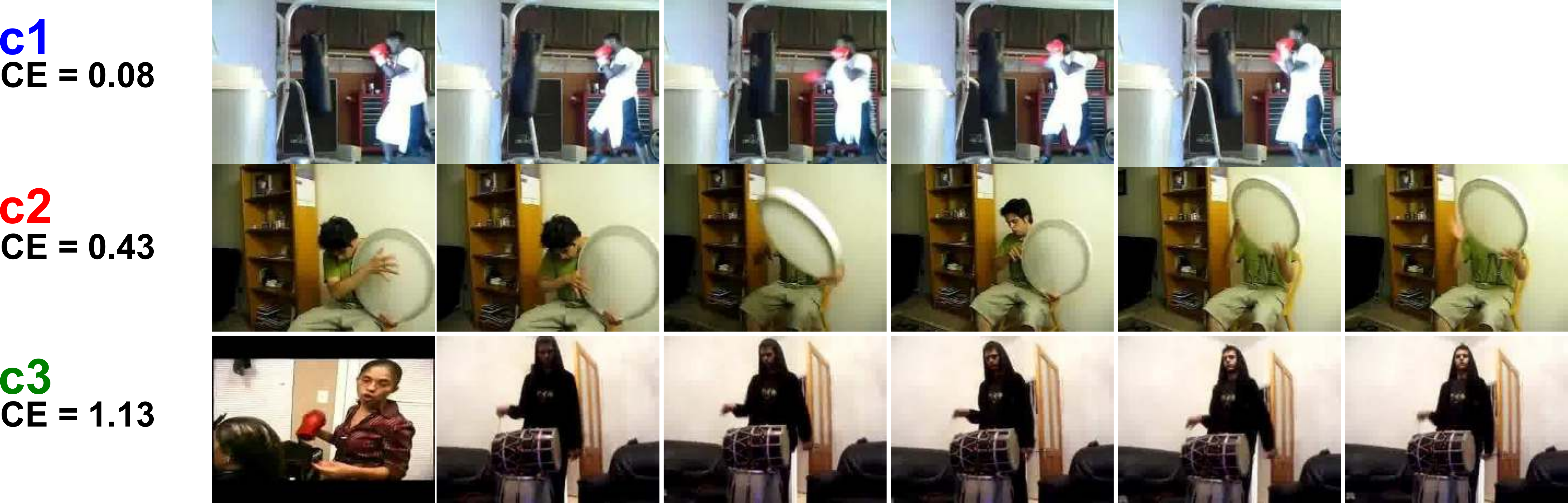}
\label{ucf_kmeans_l1}
}
\subfigure[AlexNet trained with $\mathcal{L}_q$]{
\includegraphics[width=0.9\linewidth]{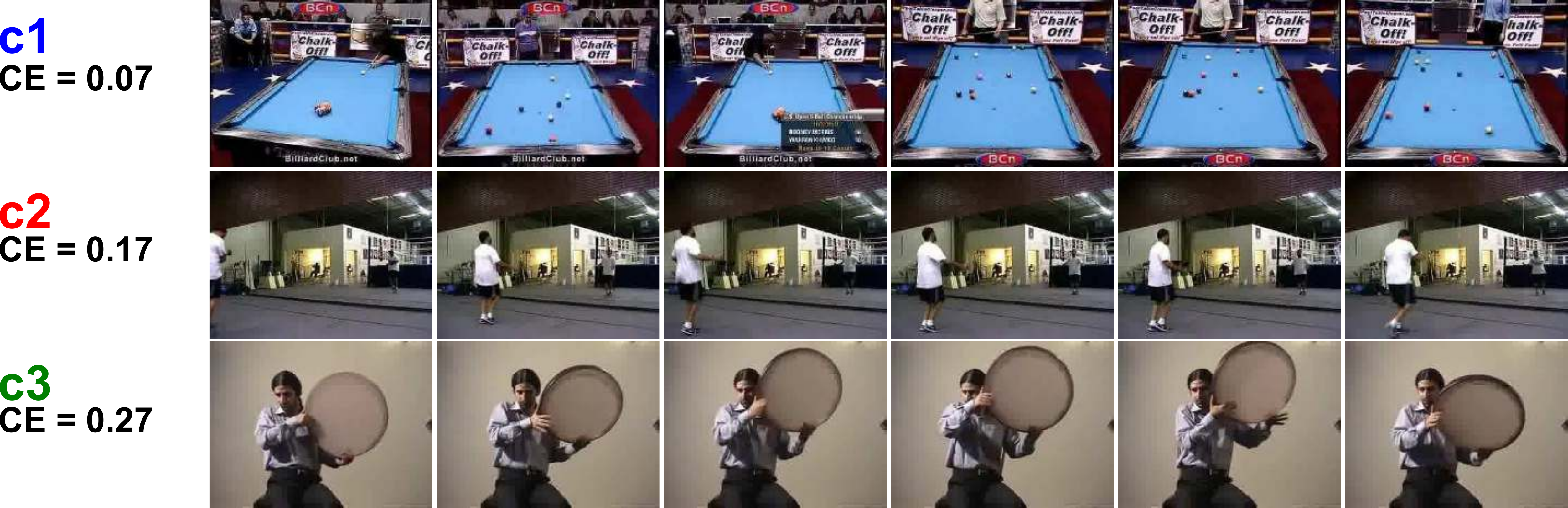}
\label{ucf_kmeans_l2}
}
\centering
\subfigure[Supervised AlexNet trained on ImageNet]{
\includegraphics[width=0.9\linewidth]{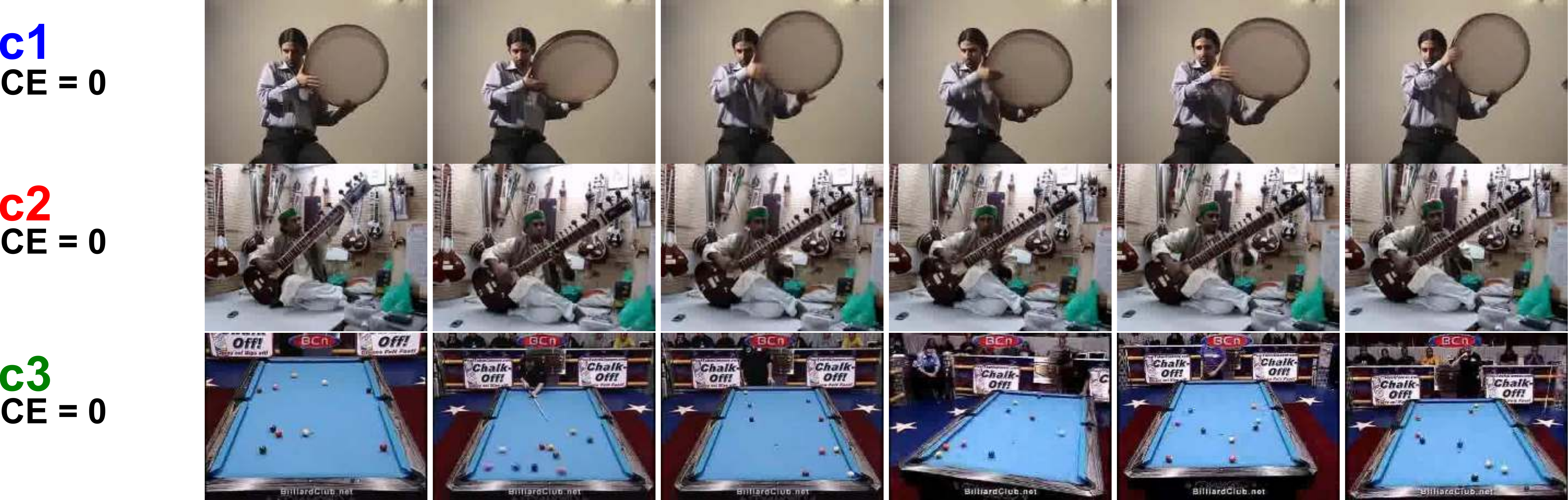}
\label{ucf_kmeans_i}
}
\caption{Qualitative results in the UCF101 dataset. 6 random images for the top three clusters discovered are shown.}
\label{fig:kmeans_qualitative_ucf}
\end{figure}

Finally, we proceed to perform a qualitative analysis of the results. Figure \ref{fig:kmeans_qualitative_ucf} shows 6 random images for the top three clusters discovered by each of the models, including the supervised solution. First, one can observe how our Quadruplet network is able to obtain a distribution very similar to that obtained by the AlexNet trained on ImageNet. Moreover, this figure also shows that our solutions are able to build clusters with an almost 0 CE without any form of supervision during the feature learning stage. 

\begin{figure}[h]
\centering
\includegraphics[width=\linewidth]{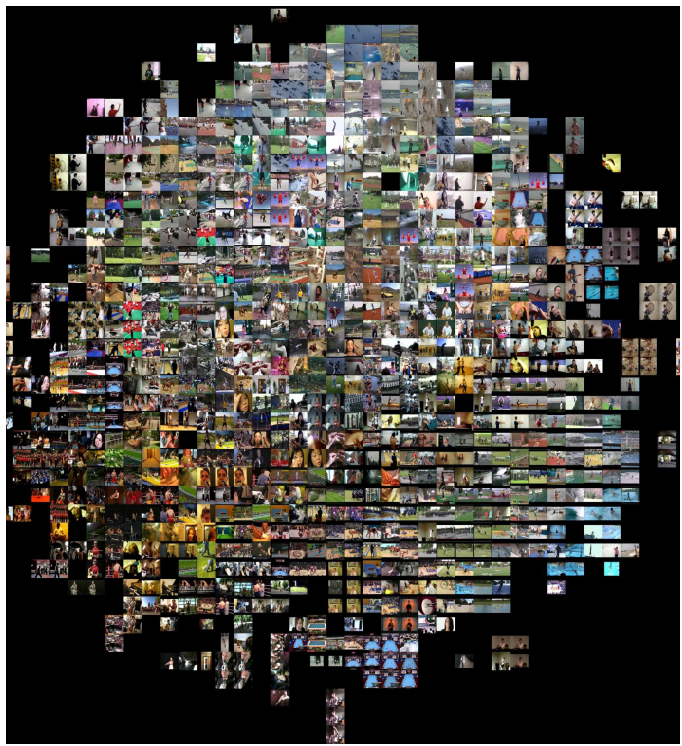}
\caption{Barnes-Hut t-SNE 2-dimensional embedding for the 4096-dimensional fc6 features learned with our Quadruplet model. Best viewed with zoom.}
\label{fig:tsne_ucf}
\end{figure}

In Figure \ref{fig:tsne_ucf} we show a 2-dimensional embedding using Barnes-Hut t-SNE \citep{Maaten2014}, for the 4096-dimensional fc6 features learned with our Quadruplet architecture. In other words, t-SNE arranges images that have a similar CNN (fc6) code nearby in the embedding. One can observe that our fully unsupervised solution is able to learn features that can actually discover groups of semantically similar images.

\subsubsection{Discovering Scenes}
\label{sec:results_scenes}

We now experimentally validate the detailed models in a different discovery task. Here we propose to use the 5-Contexts dataset for discovering scenes in personal context videos. Again, we proceed to train all our feature learning solutions, in a fully unsupervised way. $K$-means and SC based solutions are evaluated, using the features from layer fc6. According to our previous analysis in the UCF101 dataset, fc6 features perform better than fc7, being this the case for this dataset as well. Table \ref{table:ce_results_4contexts} shows a detailed analysis of the performance of our approaches, as well as a comparison with a random assignment of labels, and with a standard SFA based on a contrastive loss \citep{Hadsell2006}.

Note that the RANDOM baseline achieves a CE of $2.31$ for this dataset. This value is close to the maximum $log_{2}(5) = 2.32$, when 5 is the number of classes. First, our architectures significantly outperform the RANDOM baseline. Second, the performance of our unsupervised models is better than the achieved by the SFA approach. In a detailed comparison, we observe that also the performance of our Siamese architecture is slightly better than the one reported by the SFA model. Remember that this was not the case for the experiments in the UCF101 dataset. Now, the unsupervised models learn from egocentric videos, where the scenes and activities are largely defined by the objects that the camera wearer interacts with. Therefore, the appearance variability of these personal videos is not so high, because the different users interact with the same environments/scenes or objects (\eg the same coffee machine). These aspects clearly benefit our temporal coherency networks.

If we now analyze the results of our two architectures, we observe that the Quadruplet model is again the clear winner of the competition, reporting the lowest CE for both clustering algorithms. By applying our Quadruplet architecture, the remaining uncertainty on the true context category has been reduced from $2^{2.31} = 5.0$, for a random assignment, to $2^{1.94} = 3.8$. One more time, results reveal the benefits of using the loss $\mathcal{L}_q$, which takes advantage of jointly guaranteeing the \emph{slowness} for the video frames features, and a discriminative representation.

\begin{table}[t]
\caption{Comparison of the different unsupervised approaches using $K$-means and SC in the 5-Contexts dataset. CE (lower is better) metric is reported.}
\label{table:ce_results_4contexts}
\begin{center}
\scalebox{0.7}{
\begin{tabular}{lcc}
\toprule
Methods & $K$-means & SC\\
Baseline RANDOM & 2.31 $\pm$ 0.00 & 2.31 $\pm$ 0.00 \\
\midrule
AlexNet with $\mathcal{L}_{SFA}$ fc6  & 2.03 $\pm$ 0.01 & 1.98 $\pm$ 0.03\\
\midrule
Our AlexNet with $\mathcal{L}_s$ fc6 & 2.03 $\pm$ 0.00 & 1.98 $\pm$ 0.00\\
Our AlexNet with $\mathcal{L}_q$ fc6 & \textbf{1.94 $\pm$ 0.01} & \textbf{1.97 $\pm$ 0.00}\\
\bottomrule
\end{tabular}
}
\end{center}
\vspace{-0.5cm}
\end{table}

It is now worth to compare the performance of all these unsupervised solutions against the supervised feature learning model Alexnet \citep{Krizhevsky2012}, using ImageNet dataset. Using a  $K$-means clustering with these supervised features, the CE reported is of 0.71 $\pm$ 0.18. This excellent result can be explained by the fact we already pointed above: the scenes in these egocentric videos are defined by the objects the user interact with, therefore an AlexNet model trained for object classification should report a good performance. 

\begin{figure}[h]
\subfigure[AlexNet trained with $\mathcal{L}_s$]{
\includegraphics[width=0.9\linewidth]{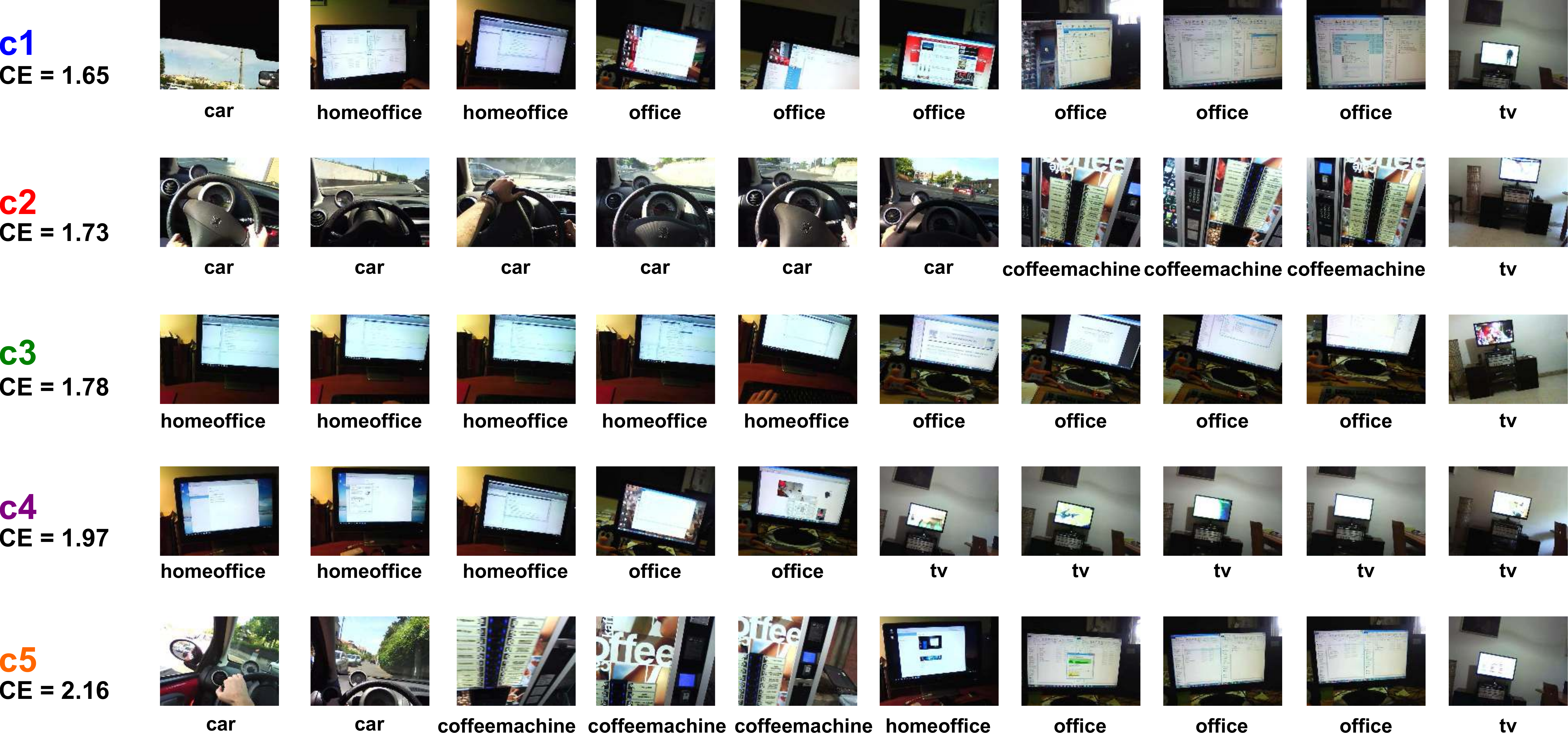}
\label{kmeans_l1}
}
\subfigure[AlexNet trained with $\mathcal{L}_q$]{
\includegraphics[width=0.9\linewidth]{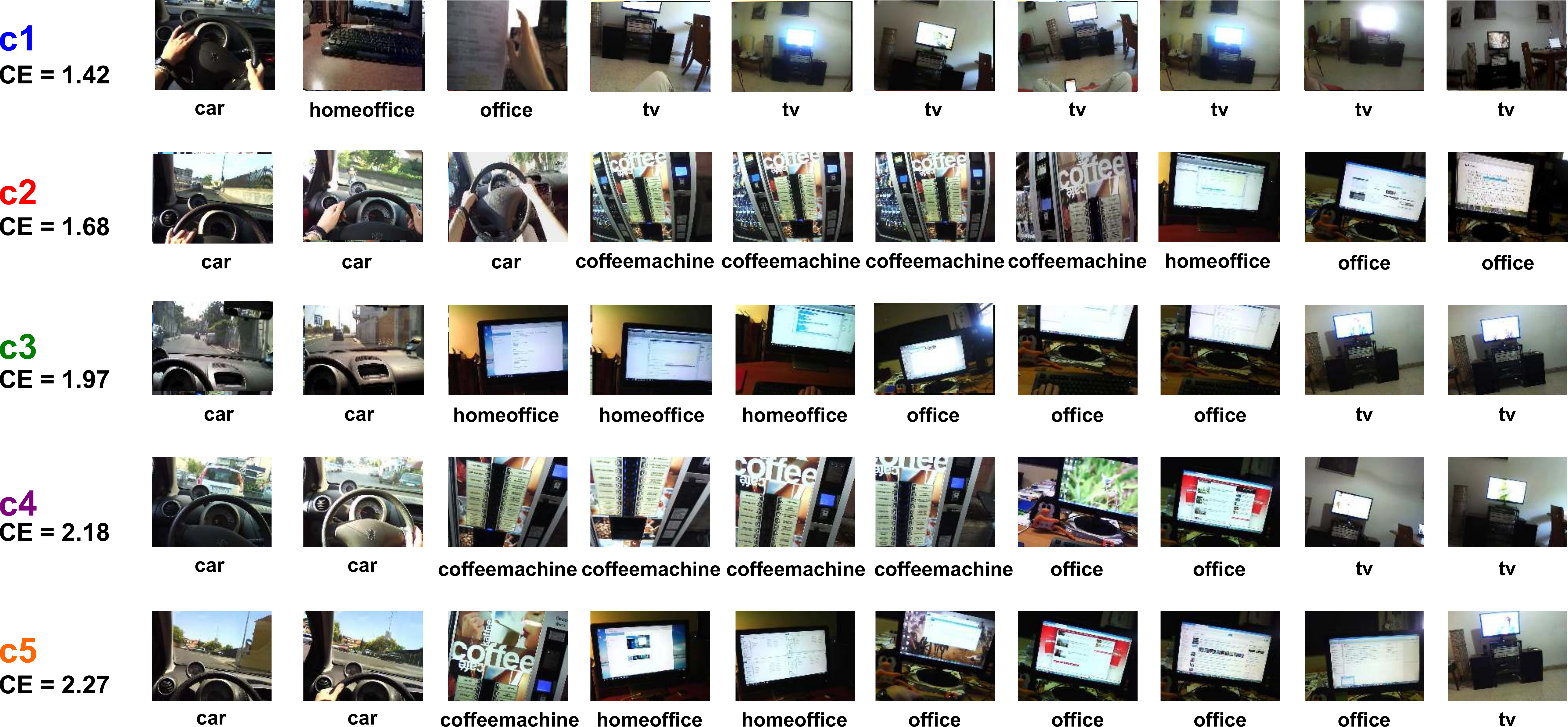}
\label{kmeans_l2}
}
\centering
\subfigure[Fully supervised AlexNet trained on ImageNet]{
\includegraphics[width=0.9\linewidth]{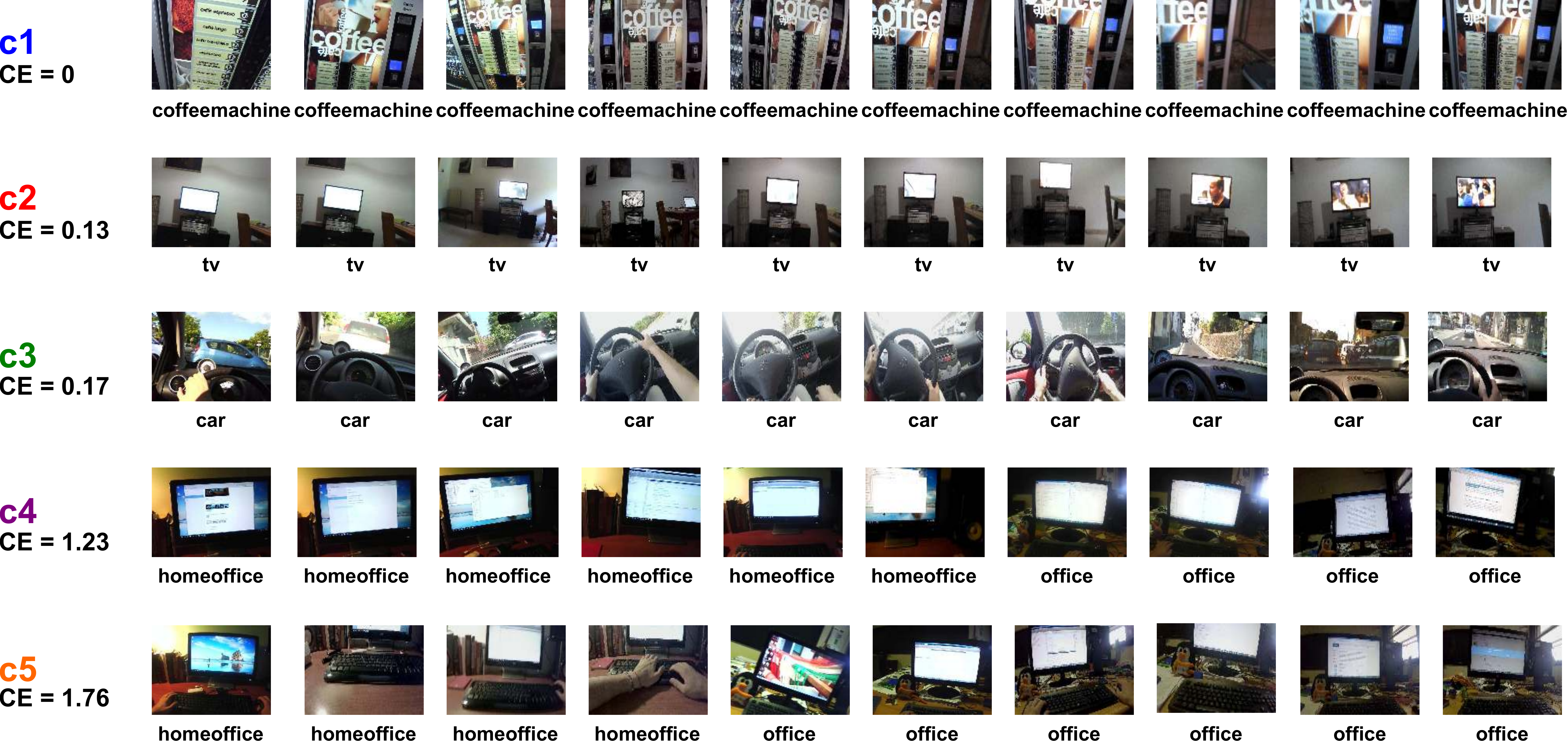}
\label{kmeans_i}
}
\caption{Qualitative results in the 5-Contexts dataset. 10 random images for each of the 5 categories discovered.}
\label{fig:kmeans_qualitative_context}
\end{figure}

We end this experiment showing qualitative results in Figure \ref{fig:kmeans_qualitative_context}. Figures \ref{kmeans_l1} and \ref{kmeans_l2} show that the main confusion for our models is for scenes \textit{home office} and \textit{office}. Overall, our approaches seem to be able to discover a finer granularity than expected, \eg joining all the video frames containing any type of screen, an object category shared across three different scenes (\textit{office}, \textit{home office} and \textit{tv}), or even including in the same cluster those images of the windshield which can be confused with a monitor too.

\subsubsection{Do our unsupervised features generalize well?}
\label{sec:transfer_experiment}

We present here an additional experiment, for which our aim is to evaluate whether our learned visual representations are able to generalize well.

The idea is simple: we propose to specifically measure the performance of the deep architecture trained with UCF101 video frames, but for discovering scenes over the test videos in the 5-Contexts dataset. We evaluate our Quadruplet Siamese AlexNet network, and use a $K$-means clustering for the discovery task. Surprisingly, we are able to report a CE of 1.80 $\pm$ 0.09. This number clearly outperforms our previously best CE (1.94). This fact reveals that our models clearly benefit from using more and more videos for learning the feature encoding. Actually, the more the heterogeneity of the videos for training, the better the capability of the features of the models to discover categories.

\section{Unlabeled video as a prior for supervised recognition}
\label{sec:supervised}

In this section we explore the benefits of initializing deep networks for a supervised learning task with the weights provided by our fully unsupervised strategies. 

Typically, the  standard recipe for obtaining a good performance on a novel recognition task using deep learning models, suggests to start initializing the weights of the networks with a pretrained model, which has been learned in a supervised way using a large scale dataset (\eg ImageNet). Second step consists in a fine-tuning process to adapt the network to the new problem or task. On the contrary, we suggest to change the initialization strategy. Instead of using the weights of a supervised model, we simply propose to initialize the models following our unsupervised learning methodology.

In this section we present a detailed experimental evaluation, where we compare these two initialization strategies for two different recognition tasks: action recognition and object classification.

\subsection{Experiments}
\subsubsection{Experimental Setup}
\label{sec:exp_setup_recognition}

Two datasets are used for the proposed tasks. Action recognition is evaluated with the PASCAL VOC 2012 dataset \citep{VOC2012}. Here we strictly follow the fine-tuning procedure detailed in \citep{Jayaraman2016}. The PASCAL VOC 2012 dataset offers 10 action categories. The images are cropped using the action annotations provided. Only 50 images (10 classes $\times$ 5 images per class) are used during the fine-tuning stage. We test on 2000 images randomly extracted from the validation set. For the object recognition experiment, we use the CIFAR-10 dataset \citep{Krizhevsky2009}. CIFAR-10 consists of 60.000 $32\times32$ color images of 10 classes, with 6000 images per category. Here, we follow the experimental setup proposed by \cite{Krizhevsky2009}: we use the 50.000 training images during the fine-tuning stage, and then we use the 10.000 testing images to test the models.

The video datasets we use to perform our fully unsupervised learning are UCF101 \citep{Soomro2012} and HMDB51 \citep{Kuehne2011}. For the former, random pairs or tuples of frames are extracted from the videos provided in the Split 1. For the HMDB51 dataset, 1000 videos clips have been randomly selected, and they have been used to extract the frames for our unsupervised feature learning.

\subsubsection{Action Recognition}
\label{action_recognition}
\begin{table}[t]
\caption{Fine-tuning results on PASCAL VOC 2012 Action dataset. Comparison using AlexNet-based CNN architectures.}
\label{table:finetuning_results}
\begin{center}
\scalebox{0.5}{
\begin{tabular}{lc}
\toprule
\multicolumn{1}{c}{Models} & Accuracy ($\%$)\\
\midrule
Baseline (Random assignment) & 9.6 \\
AlexNet Init Random & 15.10 \\
\midrule
\textbf{Our} AlexNet init. with $\mathcal{L}_s$ (UCF101 $\rightarrow$ PASCAL VOC 2012)& 16.15\\
\textbf{Our} AlexNet init. with $\mathcal{L}_q$ (UCF101 $\rightarrow$ PASCAL VOC 2012)& 18.15\\
\midrule
AlexNet init. with $\mathcal{L}_{SFA}$  (UCF101 $\rightarrow$ PASCAL VOC 2012) & 16.85\\
\midrule
AlexNet init. with \citep{Misra2016} (UCF101 $\rightarrow$ PASCAL VOC 2012)& 17.9\\
AlexNet init. with \citep{Wang2015a} (Trained on 8M data $\rightarrow$ PASCAL VOC 2012)& 22.5\\
\midrule
Supervised AlexNet pretrained on ImageNet &  \textbf{28.45}\\
\bottomrule
\end{tabular}
}
\end{center}
\end{table}

We start the experimental evaluation training our unsupervised models with the videos in the UCF101 dataset. We use the learned model weights to initialize an AlexNet network, before the fine-tuning for action recognition is performed. 

Table \ref{table:finetuning_results} summarizes the main results. We first report the performance of two baselines: 1) a random assignment of action categories (this is the sanity check); and 2) a random initialization of the weights of the network before the fine-tuning. Interestingly, this second baseline gets an action classification accuracy of 15\%. 

We now report the performance of our two unsupervised feature learning models, \ie the Siamese with loss $\mathcal{L}_s$  and the Quadruplet with loss $\mathcal{L}_q$. Additionally, we report the performance of a Siamese model where we implement the SFA model with a contrastive loss function \citep{Hadsell2006}. Clearly, our Quadruplet solution outperforms not only the baselines by a large margin, but also the Siamese models.

Table \ref{table:finetuning_results} also shows a comparison with state-of-the-art models \citep{Misra2016,Wang2015}. They both are able to learn deep models from unlabeled videos. For these experiments, we have used their available pre-trained models. The model of \cite{Misra2016} was trained using also the UCF101 dataset, and exactly the same videos that we use for our models. However, the model of \cite{Wang2015} uses 8 million triplets of frames extracted from 100k YouTube videos using the URLs provided by \cite{Liang2015}. We and \cite{Misra2016} use only 9537 videos. Overall, our Quadruplet network is able to improve \citep{Misra2016}, but \cite{Wang2015a} achieve the best accuracy. Recall that \cite{Wang2015a} model needs to work with tracked patches. This requires an extra tracking process which needs the computation of local features (SURF \citep{Bay2008}) and the costly Improved Trajectories \citep{Wang2013}. Our models simply learn from video frames in a completely end-to-end fashion.

Finally, Table \ref{table:finetuning_results} shows the accuracy obtained when a supervised AlexNet model, pretrained using the whole ImageNet, is used. Our best unsupervised solution is able to recover 64\% of the accuracy of the supervised solution. We consider these results promising, indicating that our models are able to learn effective visual representations for the action recognition task. Moreover, when the same videos are used for learning, we improve the state-of-the-art defined by \cite{Misra2016}.
 
We also want to evaluate the performance of our models in the the experiment described in \citep{Jayaraman2016}. The experimental setup proposed by \cite{Jayaraman2016} consists in initializing a CIFAR CNN architecture \citep{Krizhevsky2009} using unsupervised models for feature learning on 1000 videos clips randomly extracted from the HMDB51 dataset. Then, a fine-tuning for the task of action recognition in the PASCAL VOC 2012 has to be performed.

We compare here the following model parameters initialization strategies: 1) a random assignment; 2) using the weights obtained by our unsupervised feature learning models; 3) a Siamese architecture trained for SFA with a contrastive loss \citep{Hadsell2006}; 4) the approach of \cite{Jayaraman2016}; and 5) implementing a supervised pipeline, initially training the model to recognize the classes in the CIFAR-100 dataset \citep{Krizhevsky2009}.

Table \ref{table:finetuning_results_cifar} summarizes the main results. First, our Siamese and Quadruplet solutions outperform not only the standard random initialization baseline, but also the initialization using the SFA model \citep{Hadsell2006}. Like in all previous experiments, the Quadruplet model improves all the Siamese architectures. 

The excellent work of \cite{Jayaraman2016} reports the best performance here: it slightly improves the accuracy of the supervised pipeline (20.95\% vs. 20.22\%). The accuracy of our Quadruplet model is of 18.75\%, which means that we are able to recover 92.73\% of the accuracy of a supervised model. In fact, the supervised  model needs at least 17.400\footnote{This number has been provided by the authors of \citep{Jayaraman2016}} training images to report an accuracy of 17.35\%, which is closer to ours. We consider these numbers very promising results. Technically, our experiments show that our features, learned from unlabeled video can even surpass a standard supervised approach.

\subsubsection{Recognizing Object Categories}

\label{object_recognition}
\begin{table}[t]
\caption{Fine-tuning results on PASCAL VOC 2012 Action dataset. Comparison using CIFAR-based CNN architectures.} 
\label{table:finetuning_results_cifar}
\begin{center}
\scalebox{0.5}{
\begin{tabular}{lc}
\hline
\multicolumn{1}{c}{Models} & Accuracy ($\%$)\\
\toprule
CIFAR Init Random & 15.20 \\
\midrule
\textbf{Our} CIFAR init with $\mathcal{L}_s$ (HMDB51 $\rightarrow$ PASCAL VOC 2012)& 18.40 \\
\textbf{Our} CIFAR init with $\mathcal{L}_q$ (HMDB51 $\rightarrow$ PASCAL VOC 2012)& 18.75 \\ 
\midrule
CIFAR init with $\mathcal{L}_{SFA}$ (HMDB51 $\rightarrow$ PASCAL VOC 2012) & 16.90 \\
\midrule
CIFAR init with \cite{Jayaraman2016} (HMDB51 $\rightarrow$ PASCAL VOC 2012)& \textbf{20.95} \\
\midrule
CIFAR pretrained on 17.400 images from CIFAR-100 & 17.35 \\
CIFAR pretrained on full CIFAR-100 & 20.22 \\
\hline
\end{tabular}
}
\end{center}
\end{table}

We want to finish the experimental validation evaluating the performance of our models for the problem of image categorization. Therefore, we use here the CIFAR-10 dataset \citep{Krizhevsky2009} for image classification. Details of the experimental setup have been provided in Section \ref{sec:exp_setup_recognition}. Technically, our approach consists in initializing the CIFAR CNN architecture detailed by \cite{Krizhevsky2009}\footnote{We use the standard Krizhevsky's CIFAR network provided by Caffe \citep{jia2014caffe}}, using the model parameters learned with our unsupervised strategies for feature learning, using 1000 video clips randomly extracted from the HMDB51 dataset. Then, a fine-tuning for the task of image recognition in the CIFAR-10 dataset is performed. Note that the fine-tuning is done only with the training set of images of this dataset.

In Table \ref{table:finetuning_cifar_results} we compare the classification accuracy of three different architecture initialization strategies: 1) a random initialization of weights; 2) the approach of \cite{Wang2015a}; and 3) our best unsupervised model, \ie the Quadruplet architecture. We have implemented the ranking Euclidean loss described in \citep{Wang2015a}, using a Triplet-Siamese architecture adapted from the CIFAR network provided by Caffe \citep{jia2014caffe}. We name this approach as CIFAR init with $\mathcal{L}_{d}$ \citep{Wang2015a} in Table \ref{table:finetuning_cifar_results}. Note that all the CNN architectures in Table \ref{table:finetuning_cifar_results} are the same. We simply change the initialization strategy.

Our model outperforms the random initialization, showing that it is also feasible to learn effective visual representations for the problem of object recognition from unlabeled videos. Remarkably, training under the same conditions, our Quadruplet Siamese model slightly outperforms the learning procedure of \cite{Wang2015a}.

\begin{table}[t]
\caption{Object recognition experiment. Accuracy comparison on CIFAR-10 dataset.}
\label{table:finetuning_cifar_results}
\begin{center}
\scalebox{0.6}{
\begin{tabular}{lc}
\toprule
\multicolumn{1}{c}{Models} & Accuracy ($\%$)\\
\midrule
CIFAR Init Random & 75.1\\
\midrule
CIFAR init with $\mathcal{L}_d$ \cite{Wang2015a} (HMDB51 $\rightarrow$ CIFAR-10) & 76.1 \\
\textbf{Our} CIFAR init with $\mathcal{L}_q$ (HMDB51 $\rightarrow$ CIFAR-10) & \textbf{76.5} \\
\bottomrule
\end{tabular}
}
\end{center}
\end{table}
\section{Conclusions}
\label{sec:conclusion}

In this paper we introduce two novel CNN architectures for unsupervised feature learning from unlabeled videos. As a form of self-supervision, they jointly exploit the local temporal coherence between contiguous frames, and a global discriminative margin used to separate representations between different videos.

Using four diverse datasets, we show the impact of our methods on activity and scene discovery tasks, and on activity and object recognition problems. For the action classification problem, our experiments show that our unsupervised models can even surpass a supervised pre-training approach.

Overall, we consider promising the results reported, confirming that learning from unlabeled video can augment visual learning.

We release the code needed to reproduce the results\footnote{\url{https://github.com/gramuah/unsupervised}}.

\section{Acknowledgments}
This work is supported by project PREPEATE, with reference TEC2016-80326-R, of the Spanish Ministry of Economy, Industry and Competitiveness. We gratefully acknowledge the support of NVIDIA Corporation with the donation of a Titan X Pascal GPU used for this research. Cloud computing resources were kindly provided through a Microsoft Azure for Research Award. We want to thank the reviewers for their thorough reviews, that helped us to improve the paper. Finally, we thank Carlos Herranz-Perdiguero for his help with some experiments.
\bibliographystyle{model2-names}
\bibliography{egbib}
\end{document}